\documentclass[a4paper,conference]{IEEEtran}
\IEEEoverridecommandlockouts
\usepackage{cite}
\usepackage{amsmath,amssymb,amsfonts}
\usepackage[ruled,linesnumbered]{algorithm2e}
\SetAlgorithmName{Pseudocode}{}{}
\usepackage{graphicx}
\usepackage{textcomp}
\usepackage{xcolor}
\usepackage[top=0.705in, left=0.514in, bottom=1.605in, right=0.514in]{geometry}
\def\BibTeX{{\rm B\kern-.05em{\sc i\kern-.025em b}\kern-.08em
    T\kern-.1667em\lower.7ex\hbox{E}\kern-.125emX}}
\begin{document}

\title{Approach for Document Detection by Contours and Contrasts
\thanks{This work is partially financially supported by Russian Foundation for Basic Research, projects 17-29-03236 and 17-29-03170.}
}

\author{
\IEEEauthorblockN{
Daniil V. Tropin$^{1, 2,}$\IEEEauthorrefmark{2}, 
Sergey A. Ilyuhin$^{1, 2,}$\IEEEauthorrefmark{2}, 
Dmitry P. Nikolaev$^{1, 3,}$\IEEEauthorrefmark{4} and 
Vladimir V. Arlazarov$^{1, 4,}$\IEEEauthorrefmark{2}}

\IEEEauthorblockA{$^1$ Smart Engines Service LLC, Moscow, Russia}

\IEEEauthorblockA{$^2$ Moscow Institute of Physics and Technology (National Research University), Dolgoprudny, Russia }

\IEEEauthorblockA{$^3$ Institute for Information Transmission Problems (Kharkevich Institute) RAS, Moscow, Russia}

\IEEEauthorblockA{$^4$ Federal Research Center “Computer Science and Control” RAS, Moscow, Russia}

\IEEEauthorblockA{\IEEEauthorrefmark{2} Email: \{daniil\_tropin, ilyuhinsa, vva\}@smartengines.com}

\IEEEauthorblockA{\IEEEauthorrefmark{4} Email: dimonstr@iitp.ru}
}

\maketitle

\begin{abstract}
This paper considers arbitrary document detection performed on a mobile device.
The classical contour-based approach often fails in cases featuring occlusion, complex background, or blur.
The region-based approach, which relies on the contrast between object and background, does not have application limitations, however, its known implementations are highly resource-consuming. 
We propose a modification of the contour-based method, in which the competing contour location hypotheses are ranked according to the contrast between the areas inside and outside the border.
In the experiments, such modification allows for the decrease of alternatives ordering errors by 40\% and the decrease of the overall detection errors by 10\%.
The proposed method provides unmatched state-of-the-art performance on the open MIDV-500 dataset, and it demonstrates results comparable with state-of-the-art performance on the SmartDoc dataset.

\end{abstract}


\section{Introduction}
\label{sec:intro}

\begin{figure}[htb]

\footnotesize
\begin{minipage}[b]{0.48\linewidth}
  \centering
  \centerline{\includegraphics[width=\textwidth]{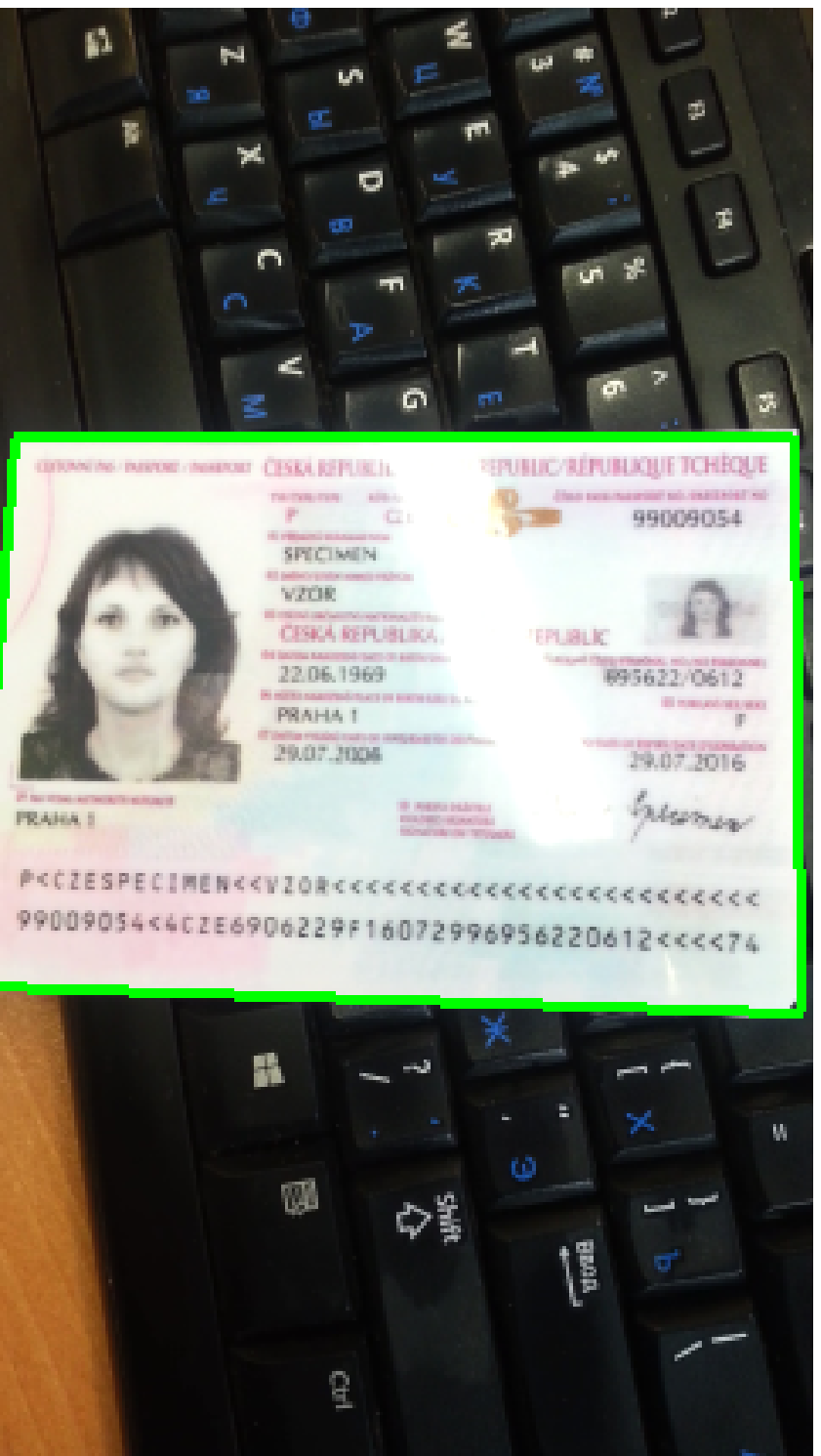}}
%  \vspace{1.5cm}
  \centerline{(a)}%\medskip
\end{minipage}
\hfill
\begin{minipage}[b]{0.48\linewidth}
  \centering
  \centerline{\includegraphics[width=\textwidth]{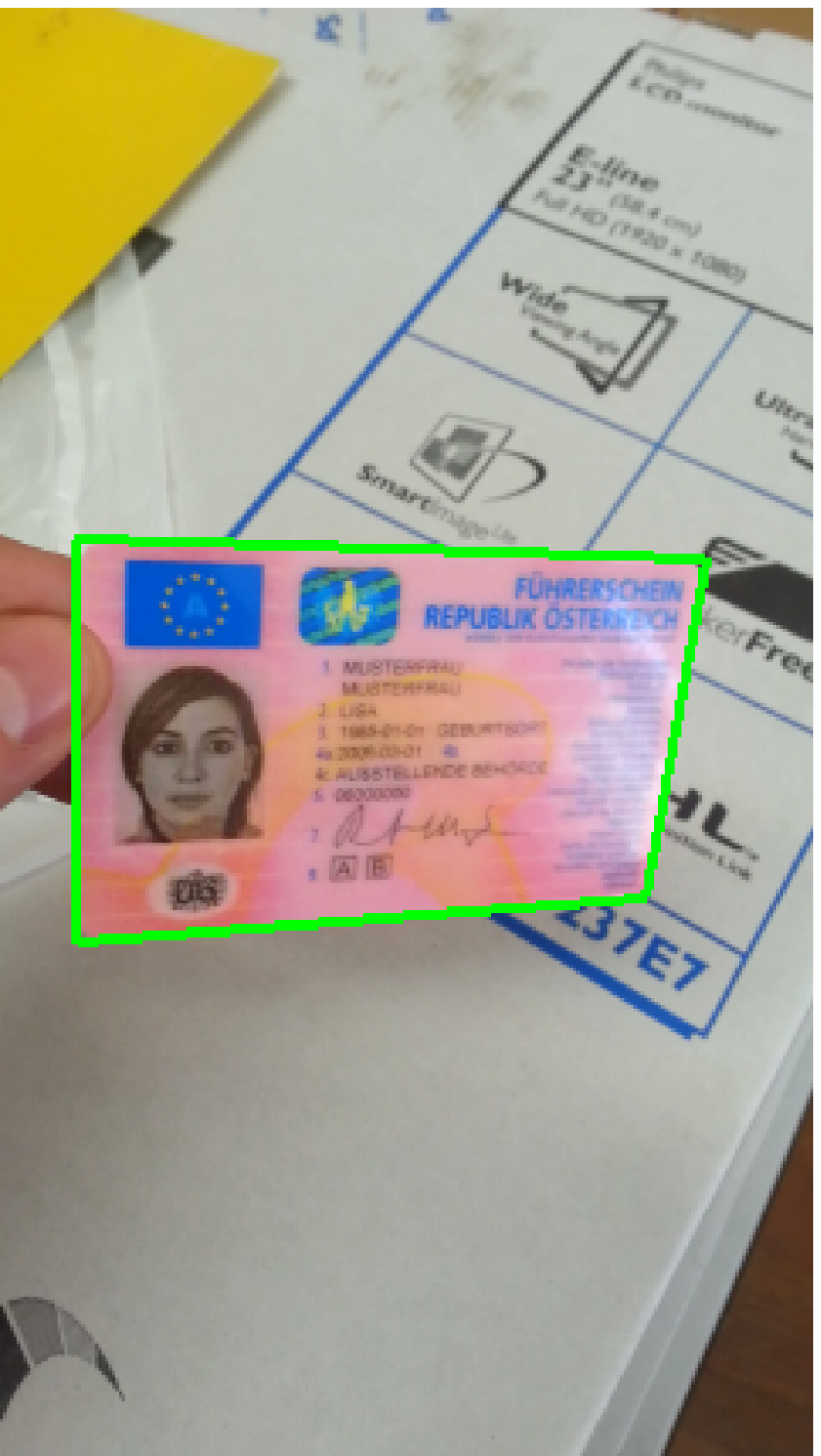}}
%  \vspace{1.5cm}
  \centerline{(b)}%\medskip
\end{minipage}

\caption{Sample images from MIDV-500 \cite{arlazarov2019midv} with highlighted documents to be detected.}
\label{fig:problem}

\end{figure}

Quadrilateral detection problem (Fig. \ref{fig:problem}) quite often takes place in computer vision tasks.
This is not surprising since rectangular objects are widespread in the urban scenes, and four points are enough to form homography basis for a planar part of the scene, and perform a local rectification. 
Thus, quadrilateral detection is used, for example, in automatic processing of documents~\cite{attivissimo2019automatic}, signboards~\cite{tam2003quadrilateral},  whiteboards~\cite{zhang2007whiteboard}, vehicle license plates~\cite{duan2004combining}, road signs~\cite{wenzel2017corners}, paintings~\cite{skoryukina20192d}, beds in hospital~\cite{kittipanya2012bed}, and etc.
Another application is an estimation of the observer orientation in scene's coordinate system specified by a rectangular object, for example, by a truck body \cite{sark2019artractor} or by corners of an evacuation plan \cite{peter2011using}.

In this paper, we consider the quadrilateral detection problem for flat document detection.
%In this paper, we solve the problem of quadrilateral document borders detection.
We assume that the image is known to contain only one document, this document has an unknown internal structure (for instance, this is the case for bank cards recognition).
Also, we suppose that there is no a priori information about camera intrinsic parameters (for example, back focal distance).

The classical contour-based approach to this problem is based on the following steps: edge detection, borders candidate selection, quadrilaterals candidate construction, and, finally, ranking of these candidates (alternatives) to choose the best one \cite{zhang2007whiteboard, liu2018dynamic, skoryukina2015real, lampert2005oblivious, hirzer2008marker}.
The key assumption of this approach is that borders of the target object form consistent and unambiguous strong poly-line contour on the image.
In other words, contour-based algorithms usually have difficulties when sought contour does not entirely fit the image or is partially occluded, when contrast between object and background is low, in the presence of strong lines crossing object or background, in poor lighting conditions, or when the image is blurred.

Another well-known approach employs known object structure: feature points \cite{awal2017complex}, sets of line segments \cite{fan2016detection}, or maps of local contrasts \cite{usilin2010visual, bohush2019video}.
These methods are hardly applicable in case of the absence of a priori information about object structure.

%Будем рассматривать именно этот случай в данной работе.
%Случай, когда внутренняя структура документа неизвестна, встречается, например, при распознавании банковских карт (карты разных банков имеют свой специфический дизайн).
Given the lack of knowledge about the object structure, the alternative contour detection approach is region-based.
The main difference from the contour-based approach is that region-based method relies on areal contrast between object and background rather than differential characteristics of the border.
Region-based algorithms were proposed in \cite{puybareau2018real} and \cite{ngoc2019document}, combining region-based segmentation and shape analysis.
Approach considered in \cite{puybareau2018real} is quite elegant.
The image is segmented by watersheds of filtered color gradients, and then the contours of segments are analyzed as candidates for sought quadrilateral shapes.
Though very fast, this algorithm does not perform well, and later, the same authors proposed much slower but more precise algorithm \cite{ngoc2019document}. 
Interestingly, the paper \cite{leal2016smartphone} suggesting the similar approach was published earlier, but it did not demonstrate performance of same good quality. 

The different strategy is proposed in \cite{attivissimo2019automatic}: a hypothetical position of the document is optimized in such a way that the color distributions in the internal and external areas of the quadrilateral would differ the most.
The advantage of contrast-based approach is robustness in cases featuring background lines and image blurring.
However, the computational complexity of such algorithms is usually high and their usage on mobile processors is limited.

In addition to contour and region approaches to the detection of quadrilaterals with unknown internal structure, there are works in which the quadrilateral is detected via set of vertices \cite {wenzel2017corners, javed2017real, zhu2019coarse}. 
%Слабой стороной такого подхода является предположение о том, что вершины должны быть видимы, следовательно, не следует ожидать высокого качества, когда они либо заходят за пределы кадра, либо заслонены.
The disadvantage of this approach is the assumption of the visible vertices. Therefore, high quality should not be expected when the vertices are obscured or located outside the frame.
%В работе \cite{zhukovsky2017segments} используется комбинация детекторов углов и контуров, что делает метод более устойчивым.
In \cite{zhukovsky2017segments}, a combination of the contours and the vertices is used. 
%Но и по отношению к таким алгоритмам критика контурных методов остается в силе.
And despite the fact that such combination makes the system more robust, the contour-based drawbacks are still present.

The problem of document detection can be reduced not only to the quadrilateral detection, but also to a semantic segmentation problem. 
This strategy was used in aforementioned region-based algorithms \cite{ngoc2019document, leal2016smartphone} and in systems \cite{junior2020fast, sheshkus2020houghencoder} based on U-Net \cite{castelblanco2020machine} architecture network.

% needs to be checked

To solve the problem, a combination of contour and region-based approaches is proposed in this paper.
Its main idea is that competing contour location hypotheses are ranked according to the contrast between the areas inside and outside the border.
We demonstrate that the usage of both approaches in the quadrilateral scoring reduces the ranking errors by 40\%.

The rest of the paper consists of three main sections: in section \ref{sec:ranking}, the ranking problem statement is discussed, then, in section \ref{sec:algo}, our algorithm is proposed, and section \ref{sec:experiments} describes the experimental evaluation.

\section{The ranking problem statement}
\label{sec:ranking}
%Пусть есть изображение $I$, множество альтернатив -- четырехугольников $\{q\}$ мощность которого равняется $N$, истинный четырехугольник $m$, задающий образ прямоугольного объекта на изображении $I$.
Let us consider an image $I$, a set of quadrilaterals $\{q\}_{i=0}^N$ (the details on quadrilaterals baseline set generations are given in paragraph \ref{subsec:quad_generation}), and the ground truth quadrilateral $m$ which describes the position of an object in the image.
%Требуется задать функцию $F(q, I)$ таким образом, чтобы альтернатива с наибольшей оценкой $q^* = \operatorname*{arg\,max}_q {F(q, I)}$ удовлетворяла истинному четырехугольнику $m$ по бинарной метрике качества $L(q, m)$. 
It is required to define a function $F(q, I)$ in such way, that the quadrilateral with the highest score $q^* = \operatorname*{arg\,max}_{i=0}^N {F(q_i, I)}$ fits the ground truth quadrilateral $m$ according to a binary quality metric $L(q, m)$.
%В случае корректной альтернативы $L=1$, иначе $L=0$.
If the tested quadrilateral $q$ is correct, then $L(q,m)=1$, otherwise $L(q,m)=0$.

\section{Proposed algorithm}
\label{sec:algo}
%Как было сказано выше, в функции $F$ предлагается учитывать, как контурные характеристики альтернативы, так и областные, а именно, будем искать функцию $F$ вида:
Function $F$ should take into account both contour and contrast features of quadrilateral $q$.
We search this function in the following form:
\begin{equation}
\label{eq:comb}
  F(q,I)=kC(q, I) + R(q, I),
\end{equation}
%где $C$ - контурная оценка, которая будет описана в разделе \ref{subsec:contour}, $R$ - областная оценка (раздел \ref{subsec:region}), $k$ коэффициент комбинирования (раздел \ref{subsec:k_optimization}). 
where $C$ is a contour-based score, which is described in subsection \ref{subsec:contour}, $R$ is a contrast-based score (subsec. \ref{subsec:region}), $k$ is a combination coefficient (subsec. \ref{subsec:k_optimization}). 

%\textcolor{red}{As long as the main topic of this paper is about developing score function $F$ of an alternative $q$ on the image $I$, description of a method acquiring that alternatives $\{q\}$ as well as the quality metric description will be discussed in sections \ref{subsec:quads_formation} and \ref{subsec:evaluation} respectively.}

\begin{figure*}[!ht]
    \centering
    \includegraphics[width=1.\textwidth]{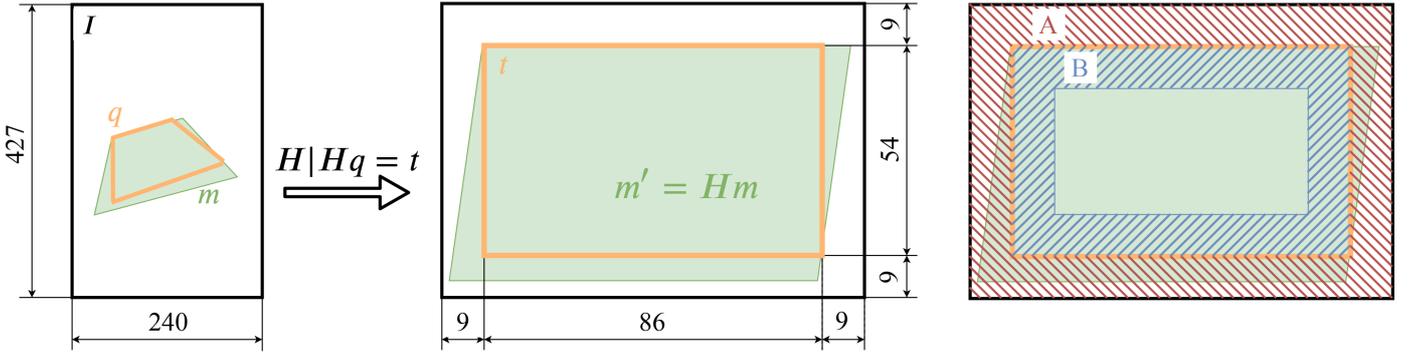}
    \caption{Scheme of obtaining the external $A$ and internal $B$ regions of the $q$.}
    \label{fig:reordAB}
\end{figure*}

\subsection{Contour-based score}
\label{subsec:contour} 

%В работах, основанных на классическом подходе, альтернатива $q$, состоящая из четырех сторон $\{b_i\}_{i=1}^4$, оценивается на основе границ типа край вдоль ее сторон.
In classical contour-based algorithms quadrilateral $q$, composed of four borders $B(q)=\{b_i\}_{i=1}^4$, is usually estimated via edges (for example, Canny edges \cite{canny1986computational}). 
%Например, в \cite{zhang2007whiteboard} альтернатива оценивается исходя из непрерывности (аналог consistency в английском) $c$ границ вдоль сторон: $\operatorname*\sum_{b} c(b)$.
For instance, in \cite{zhang2007whiteboard}, a consistency score $\operatorname*\sum_{b \in B(q)} c(b)$ of all edges is calculated, where the consistency $c$ is a ratio of non-zero edges along the border $b$.
%Непрерывность одной стороны четырехугольника вычисляется как доля точек на контуре, имеющих совпадающее со стороной направление. 
%В работе \cite{liu2018dynamic} $F$ вычисляется как $min_{b} c(b)$.
In paper \cite{liu2018dynamic}, the function $F$ is estimated as $\min_{b \in B(q)} c(b)$.
%В работе \cite{skoryukina2015real} функция оценки четырехугольника базируется на использовании интенсивности границ вдоль стороны $w$ и штрафе за выход контура за пределы четырехугольника: ${\sum_b w(b)} - \sum_b w(p_b)$, где $w(p_b)$ это штраф равный суммарной интенсивности границ, находящихся на прямой вдоль стороны $b$ на некотором расстоянии $s$.

In our work, the contour score of quadrilateral $q$ is based on the edges along the lines forming a quadrilateral.
It consists of the reward and penalty~(\ref{eq:contour_score}).
The reward is calculated based on the intensity $w$ of the edges along four borders $b \in B(q)$ of the alternative $q$, and the sum of consistencies $c$ of these borders.
The penalty is calculated as the total intensity $w'$ of the edges along the intervals which belong to the lines formed by the borders of the alternative (the intervals are located outside the border $b$ and begin at quadrilateral's corners, the length of each of these 8 intervals is 10 pixels).
\begin{equation}
\label{eq:contour_score}
    C(q) = \frac{\sum_{b \in B(q)}w(b)}{1 + \sum_{b \in B(q)}(1 - c(b)) } - \sum_{b \in B(q)} w'(b).
\end{equation}

\subsection{Contrast-based score}
\label{subsec:region}

%Оценка соответствия внутренней области альтернативы образу прямоугольного объекта с неизвестным заполнением возможна только при учете предположения о том, что заполнение объекта отличается от фона. 
The evaluation of the match between the inner region of the alternative and the internal structure of the target object (in the lack of knowledge about the object structure) is feasible only under the assumption that the texture of the object is different from the background.
%Тогда степень различия двух областей можно использовать для оценки правдоподобия альтернативы.
The degree of difference between two regions may be used for quadrilateral alternatives scoring.
%Для оценки степени различия двух областей в \cite{ngoc2019document} используется комбинация псевдо-расстояния Даху \cite{geraud2017introducing} и расстояния Хи-квадрат.
To assess the degree of difference between two regions, for example, intensity \cite{bobkov2006matching}, color \cite{attivissimo2019automatic}, or both characteristics \cite{ngoc2019document} of the image may be used.
%Authors of \cite{ngoc2019document} take into account both: they use intensity in Dahu pseudo-distance and and color in Chi-square distance.

In our case, the contrast score of $q$ is based on color difference, measured by the Chi-square distance, which was used in \cite{ngoc2019document}.
For this, two regions were obtained: external region $A = \{a\}_{j=1}^{n_A}$, and internal $B = \{b\}_{j=1}^{n_B}$ (Fig.~\ref{fig:reordAB}).
Then, two histograms, $H^A$ and $H^B$, computed on the quantized colors of all pixels in the regions $A$ and $B$ were obtained and normalized.
The final score was calculated via two histograms as:
\begin{equation}
\label{eq:region_score}
    R(q) = \sum_{j=1}^{N_b} {\frac{(H^A(j) - H^B(j))^2}{H^A(j) + H^B(j)}}
\end{equation}
where $N_b$ is a number of bins in each histogram.

\subsection{Optimization of the combination coefficient}
\label{subsec:k_optimization}

%Коэффициент комбинирования двух оценок оптимизировался исходя из обучающих данных.
The combination coefficient of two score functions was optimized using the training dataset.
%Пусть используемый для обучения датасет состоит из $M$ изображений, для каждого из которых помимо истинного четырехугольника $m$, известны $N$ альтернатив а также их контурная и областная оценки.
Let the training dataset consist of M samples.
Each sample features the ground truth quadrilateral $m$; and for each sample $N$ alternatives $\{q\}$ as well as their contour-based (\ref{eq:contour_score}) and contrast-based (\ref{eq:region_score}) scores are calculated. 
%Требуется определить такой коэффициент комбинирования $k$ при котором число успешных детекций четырехугольника по метрике $L$ максимально: $k^* = \operatorname*{arg\,max}_k \sum_{i=0}^M{L(q_i^*, m_i)}$.
We need to find the combination coefficient $k$ which allows for the largest number of the successful quadrilateral detection instances according to $L$: 
\begin{equation}
    k^* = \operatorname*{arg\,max}_k \sum_{i=0}^M{L(q_i^*, m_i)}.
\end{equation}

A solution to this problem is equivalent to finding all intervals on a line with maximum number of segment intersections.

In further experiments we used $k$, which was optimized using all images of MIDV-500 \cite{arlazarov2019midv} dataset.

\section{Experiments}
\label{sec:experiments}

\subsection{Baseline quadrilaterals generation}
\label{subsec:quad_generation}
%В качестве базовой реализации контурного подхода в задаче поиска четырехугольника была взята работа \cite{skoryukina2015real}.
For baseline set of quadrilaterals $\{q\}_{i=0}^N$ generation we used the algorithm derived from the classical contour-based approach~\cite{skoryukina2015real}.
%В ней предполагалось, что стороны образа прямоугольного объекта находится в известных областях. 

Let the image be in portrait orientation.
In \cite{skoryukina2015real}, it was assumed that the objects' borders in the image are located in known regions of interest.
%В данной работе были использованы более мягкие ограничения на допустимые положения четырехугольника на изображении -- противоположные стороны либо горизонтальные, либо вертикальные.
In our case, the weaker restrictions on the quadrilateral position in the image were used: the opposite sides are either "mostly horizontal" (the range of tangent is restricted to $[-1;1]$) or "mostly vertical"~(assumption~*).
%Учет этого предположения происходит на всех этапах получения четырехугольков, поэтому уже при выделении границ были получены две карты границ: вертикальная и горизонтальная. 
This assumption is taken into account for all stages of quadrilaterals calculation. 
Edge extraction results in two edge maps: with "mostly horizontal" and "mostly vertical" edges respectively.
%Реализация вычисления карты границ подробно описана в \cite{skoryukina2015real}.
%Затем каждая карта границ размывается с помощью гаусса в направлении градиента и для каждой карты границ вычисляется ее FHT образ и на каждой части выбирается по 15 локальных максимумов (помарка: при поиске линий вдоль длинной стороны карта границ разбивается на три равные части, для каждой из которых вычисляется FHT образ и на каждой выбирается по 15 наиболее ярких точек). 
Then, each edge map is blurred with the Gaussian filter along the direction of the gradient.
To improve short sides detection quality, the vertical edge map is divided into three equal parts by horizontal cuts prior to strongest lines search.
To search line candidates via edge maps, the Fast Hough Transform (FHT) \cite{brady1998fast} was applied.
As a result of the line search, 15 extreme points for the horizontal FHT image, and 45 for the vertical FHT image (15 from each part) were obtained.
%В итоге, имея 15 прямых вдоль короткой стороны и 45 вдоль длинной, запускается перебор четверок прямых: пар горизонтальных и вертикальных прямых. 
Next, a brute-force search of two pairs of lines (vertical and horizontal) was performed.
%В работе \cite{skoryukina2015real} при переборе четырехугольков проверяются геометрические критерии, однако в данной работе они не были использованы.
In \cite{skoryukina2015real}, while searching for the best image of the rectangular object, its geometry was checked. In our work, however, the geometry tests were not used, since our study is primarily focused on contour and contrast features analysis.
%Это было сделано с целью повышения чистоты эксперимента над контурной оценкой четырехугольков.
%Далее альтернативы упорядочивались по убыванию контурной оценки (\ref{eq:contour_score}) и фильтровались те, которые не попали в $N$ лучших.
Finally, all quadrilaterals were sorted in the descending order of the contour-based score (\ref{eq:contour_score}), and top $N$ were selected. 

\subsection{Datasets and evaluation}
\label{subsec:evaluation}

All of our quality metrics were based on Jaccard index, which was used in SmartDoc challenge \cite{burie2015icdar2015}.
As it was mentioned above, $q$ is the found (or predicted) quadrilateral, $m$ is a ground truth image of a rectangle object, and $t$ is a template of the rectangle object.
Let $H'$ represent a homography such that $H'm = t$.
Then, Jaccard index between predicted and ground truth quadrilaterals is calculated as follows: 
\begin{equation}
JI(q, m, t) = \frac{\mathrm{area}(q'\cap t)}{\mathrm{area}(q'\cup t)},
\label{eq:JI}
\end{equation}
where $q' = H'q$.

To check whether the predicted quadrilateral $q$ is correct or incorrect, quality metric $L(q, m, t)$ was used.
\begin{equation}
  L(q, m, t) = \begin{cases} 1, & \mbox{} JI(q, m, t) \leq \gamma \\ 0, & \mbox{} JI(q, m, t) > \gamma \end{cases},
  \label{eq:correct}
\end{equation}
%где $P=2(w+h)$ -- периметр шаблона, $\gamma$ -- пороговый коэффициент.
where $\gamma$ is the threshold coefficient, which was set to 0.945.

\begin{figure*}[htb]

\footnotesize
\begin{minipage}[b]{0.23\linewidth}
  \centering
  \centerline{\includegraphics[width=\textwidth]{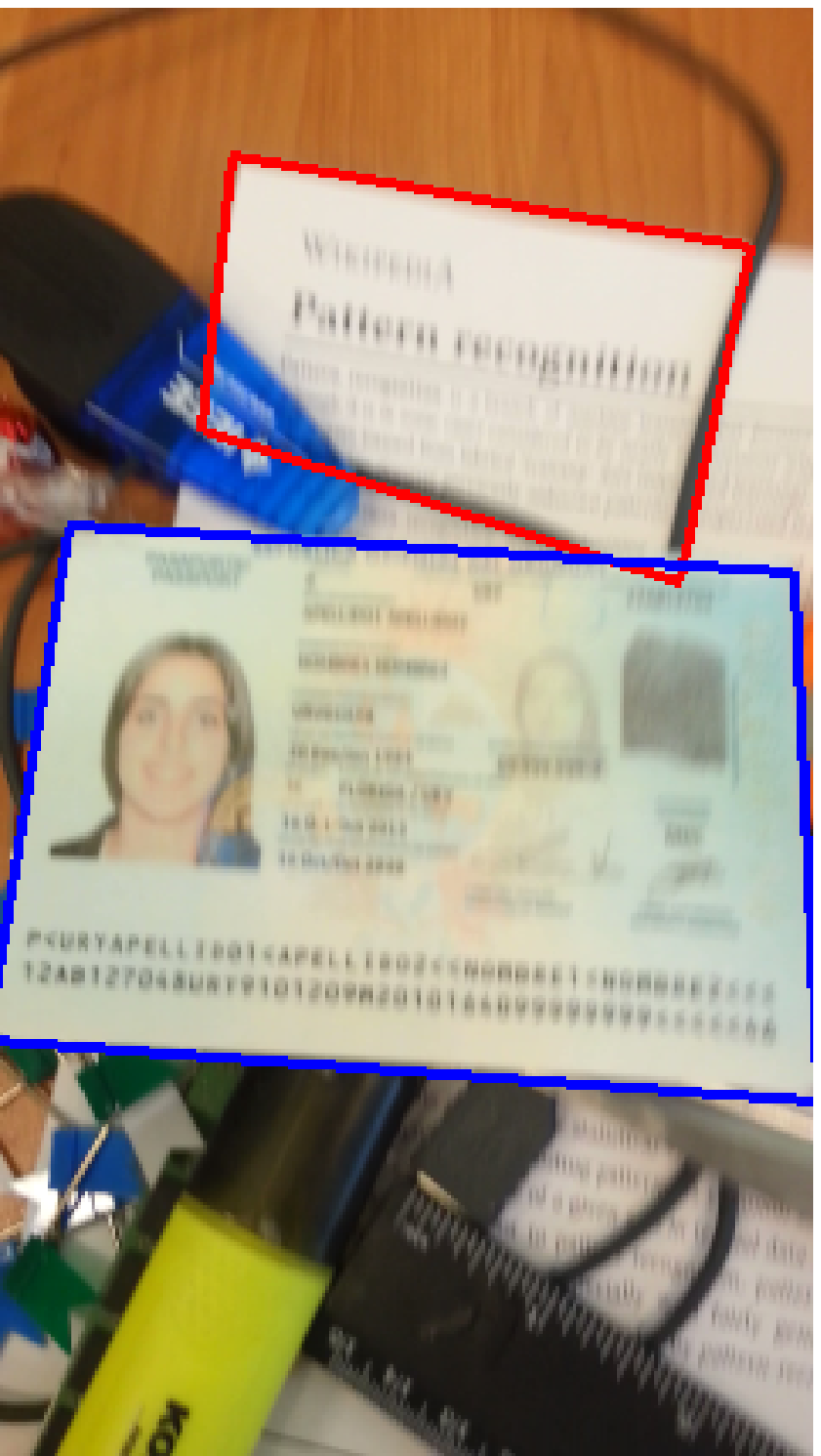}}
%  \vspace{1.5cm}
  \centerline{(a)}%\medskip
\end{minipage}
\hfill
\begin{minipage}[b]{0.23\linewidth}
  \centering
  \centerline{\includegraphics[width=\textwidth]{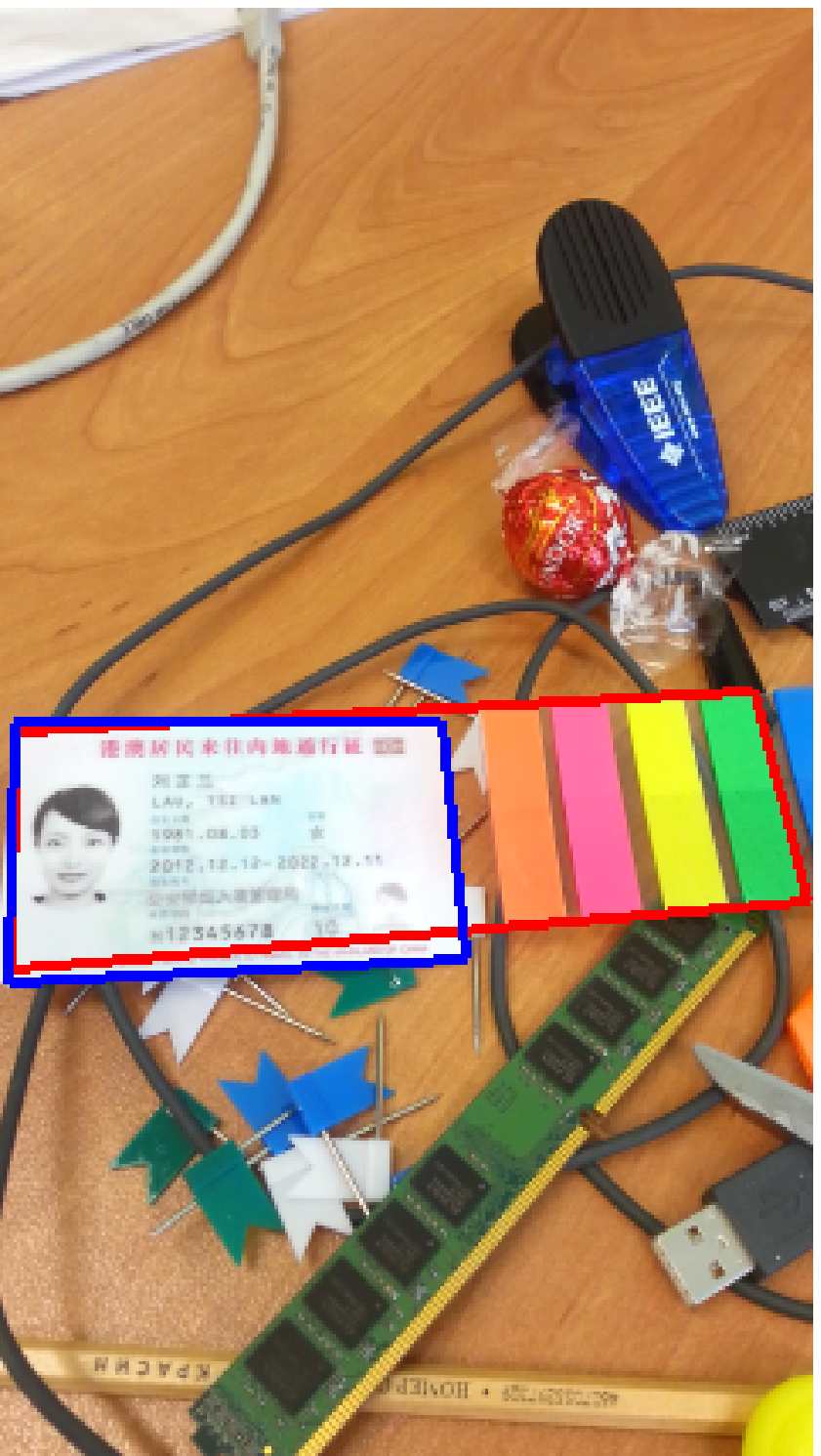}}
%  \vspace{1.5cm}
  \centerline{(b)}%\medskip
\end{minipage}
\hfill
\begin{minipage}[b]{0.23\linewidth}
  \centering
  \centerline{\includegraphics[width=\textwidth]{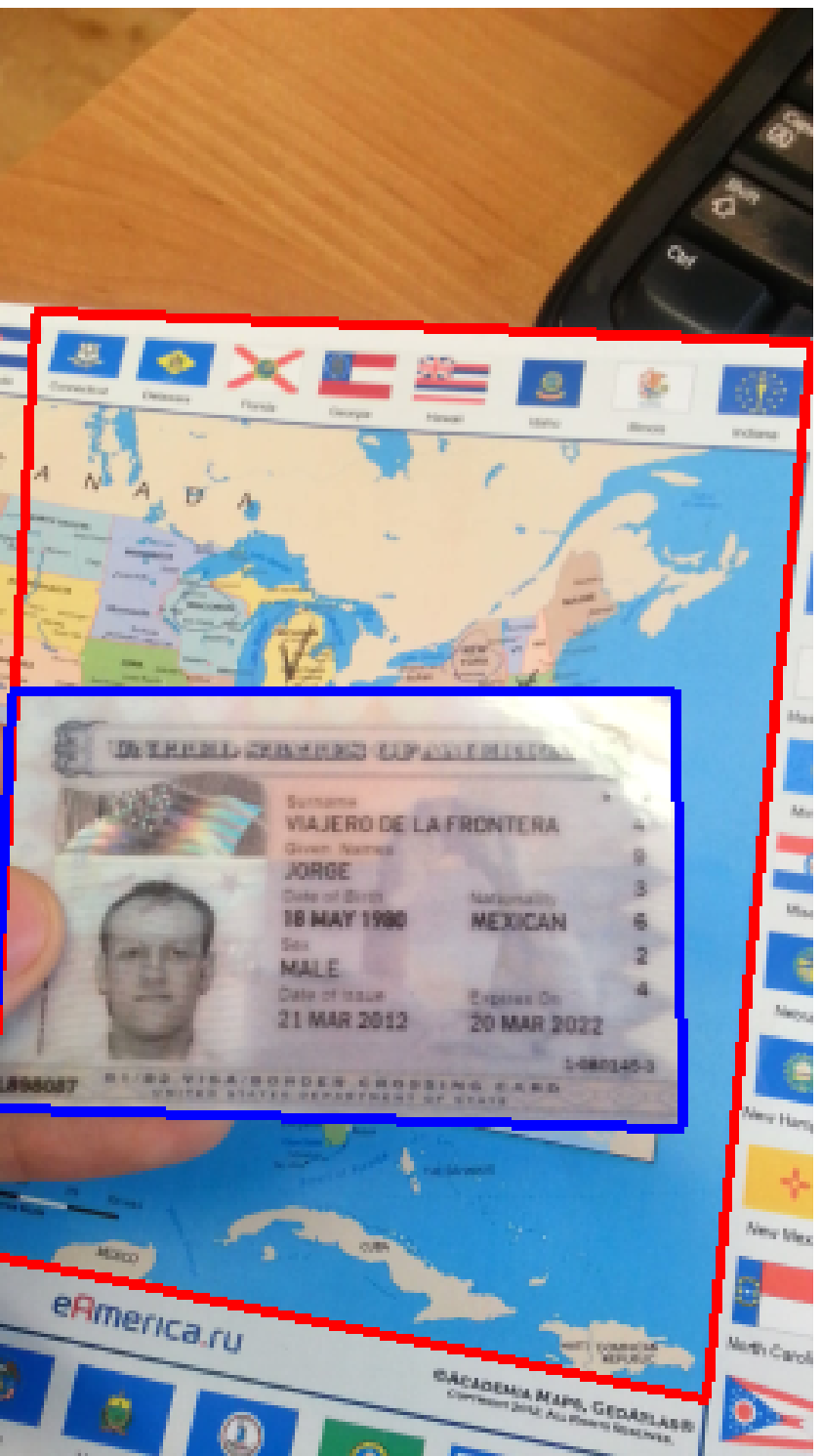}}
%  \vspace{1.5cm}
  \centerline{(c)}%\medskip
\end{minipage}
\hfill
\begin{minipage}[b]{0.23\linewidth}
  \centering
  \centerline{\includegraphics[width=\textwidth]{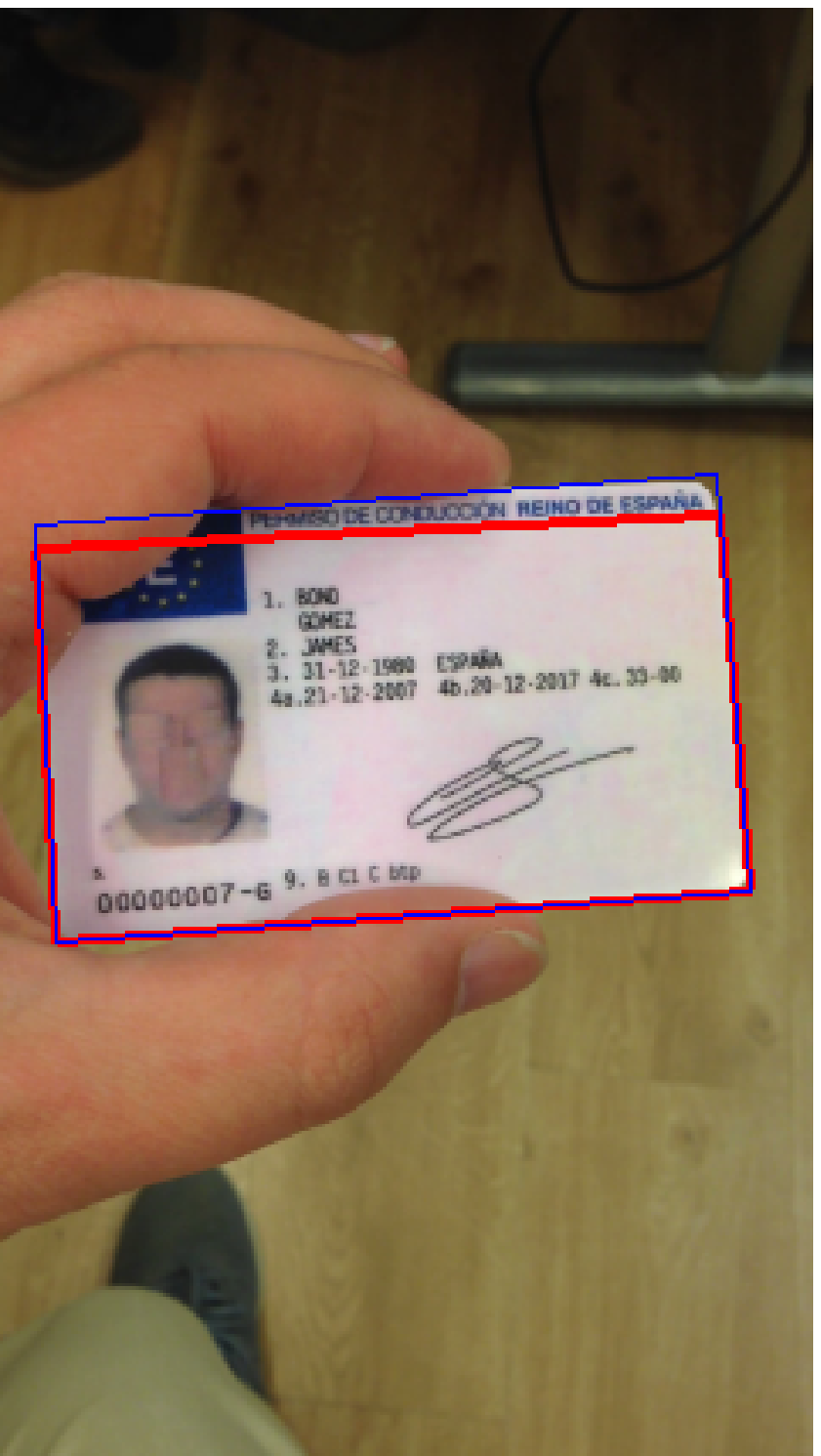}}
%  \vspace{1.5cm}
  \centerline{(d)}%\medskip
\end{minipage}

\caption{Example of the resolved errors. Red quadrilateral corresponds to the top contour alternative, blue one -- to the top combined alternative.}
\label{fig:resolved_errors}
\end{figure*}

We have tested the proposed algorithms using two open datasets: MIDV-500~\cite{arlazarov2019midv}, which contains images capturing identification documents of different types and with different backgrounds, and SmartDoc~\cite{burie2015icdar2015}, which was created specifically for A4 page detection challenge. 
The first dataset, as it was mentioned previously, contains 500 video clips (10 clips for each of the 50 unique identification documents, i.e. 15000 images in total), the ground truth quadrilateral $m$ for each frame, and information about the template sizes $t \in \{(856;540), (1050;740), (1250;880)\}$. 
We want to point out the following feature of MIDV-500: in each captured image, the document is not necessarily placed completely within the frame.
So, in our experiments, we consider several subsets of the images: first subset contains images in which all 4 vertices of the document are within the frame (9791 images), second subset contains images with at least 3 vertices within the frame (11965 images), and the third subset includes the entire MIDV-500 (all 15000 images). 

In contrast to MIDV-500, SmartDoc \cite{burie2015icdar2015} dataset guarantees that the document is placed fully within the frame.
This dataset has its own separation into subsets: it has 5 subsets with different backgrounds, from the easiest one to the most complex.

\subsection{Experiment methodology}
\label{subsec:technical}
Before we move on to the experiments, we would like to note some implementation details: 
since step-edges are scale-invariant features \cite{canny1986computational}, the original image resolution was scaled down to $240\times427$;
in equation (\ref{eq:region_score}), the histograms with 512 bins (by 8 on each of RGB channels) were used.

\begin{figure}[!ht]
    \centering
    \includegraphics[width=0.5\textwidth]{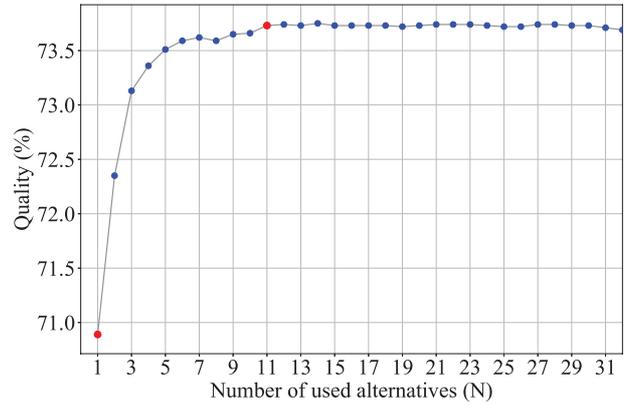}
    \caption{Dependence of the system performance quality on $N$. Two compared versions are marked in red.}
    \label{fig:n_quality}
\end{figure}

%Для запуска системы с использованием комбинированной оценки (\ref{eq:comb}) необходимо предварительно задать значение параметров $N$ и $k$.
To run the system using the combined score (\ref{eq:comb}), the values of parameters $N$ and $k$ should be defined.
%Напомним, что $N$ -- это число альтернативных четырехугольников, которые участвуют в ранжировании с помощью комбинированной оценки, $k$ -- коэффициент комбинирования контурной оценки и оценки контраста.
Let us remind that $N$ is a number of top (par. \ref{subsec:quad_generation}) quadrilateral alternatives ranked using the combined score, and $k$ is the combination coefficient~(\ref{eq:comb}).
%Для исследования системы при использовании комбинированной оценки (\ref{eq:comb}) было проведено несколько экспериментов, соответствующих разному числу рассматриваемых альтернатив $N$.
To select $N$, several experiments were performed, each featuring different number of considered alternatives (Fig.~\ref{fig:n_quality}).
%Для каждого $N$ независимо было подобрано оптимальное значение коэффициента комбинирования (sec. \ref{subsec:k_optimization}). 
For each value of $N$ independently, the optimal value of the combination coefficient $k$ was calculated (par.~\ref{subsec:k_optimization}). 
%Для оптимизации $k$ использовались данные, собранные со всего датасета MIDV500.
As it was mentioned above, all 15~000 images from the MIDV-500 dataset were used to optimize $k$.
%При использовании комбинированной оценки \ref{subsec:k_optimization} качество строго возрастает до 73\%, которое достигается при $N=15$ (Figure \ref{fig:n_quality}).
Let us consider Fig.~\ref{fig:n_quality}.
With the increase of the number of considered alternatives, the quality of the algorithm performance on the entire MIDV-500 dataset also increases up to 73.73\% until $N=11$.
%Поскольку при больших значениях $N$ качество меняется незначительно, то для сравнения с контурным подходом использовалась именно эта версия.
For large $N$, the quality changes negligibly, thus, in the following experiments, the combined score was used to rank top 11 quadrilateral alternatives.

\begin{table*}[ht]
  \caption{Comparison of two versions of the score function with error classification.}
  \label{tab:comparison}
  \centering
    \begin{tabular}{ccccc}
        \hline
        \hline
         & \begin{tabular}[c]{@{}c@{}}(i) Out of frame\end{tabular} 
         & \begin{tabular}[c]{@{}c@{}}(ii) No line\end{tabular}
         & \begin{tabular}[c]{@{}c@{}}(iii) Ranking err.\end{tabular} 
         & \begin{tabular}[c]{@{}c@{}}Total err.\end{tabular} \\ 
         \hline
         \hline
        Contour & 2850 & 660 & 854 & 4366  \\ 
        Combined & 2803 & 627 & 509 & 3941 \\ 
        %Delta+ & 11 & 15 & 82 & 108 & 3.13\% & \\ 
        %Delta- & 42 & 50 & 378 & 470 & 0.72\% & \\ 
        Improvement & 47 & 33 & 345 & 425  \\ 
        Relative improvement & 1.65\% & 5.0\% & {\bf 40.4} \% & 9.73\% \\ \hline
        \hline
    \end{tabular}
  
\end{table*}

\subsection{Results and performance analysis on MIDV-500}
\label{subsec:error_classification}
Let us start with the proposed algorithms performance quality evaluation on MIDV-500, and the error classification for each case.
Then, we demonstrate the time performance on a mobile device.

%Для тестирования эффективности предложенной комбинации контурной и областной оценки четырехугольников был использован открытый датасет MIDV500 \cite{arlazarov2019midv}, состоящий из 15000 изображений разрешения 1080x1920 различных удостоверяющих личность документов. 

%Процент успешных детекций четырехугольников при использовании только контурной оценки (\ref{eq:contour_score}) на стенде MIDV500 равен 71. 
The percentage of correctly (\ref{eq:correct}) localized quadrilaterals, while using only  contour-based score (\ref{eq:contour_score}), on the entire MIDV-500 dataset was 71\%.

All localization errors can be divided into 4 classes: 
    (i) less than 20\% of at least one of the ground truth quadrilateral's borders $b(m)$ is present within the frame,
    (ii) no lines were found in the vicinity of $b(m)$,
    (iii) ranking system error and 
    (iv) the assumption (*) is violated.
Let us note that the errors of the third class are our target.
%Распределение ошибок по классам следующее: 65\% ошибок соответствуют первому классу, 15\% второму и 20\% третьему. 
The distribution of errors by class is as follows: 65\% correspond to the first class, 15\% to the second, and 20\% to the third.
%Ошибки 4 класса были выявлены всего на 2 изображениях. 
The error percentage of the fourth class is negligible (2 errors). 
%Число ошибок каждого класса при использовании только контурной оценки приведены в первой строке Table \ref{tab:comparison}.
The number of errors for each of the first three classes if using only the contour-based score is shown in the first row of Table~\ref{tab:comparison}.

%Количество ошибок системы при использовании комбинированной оценки сократилось на 8.3\% по сравнению с общим числом ошибок базовой версии.
The number of errors if using the combined score (the second row of Table \ref{tab:comparison}) is decreased by 10\% compared to the total number of errors of the algorithm with contour-based score.
%Более 80\% всех исправленных ошибок относятся к целевому iii классу, когда правильная альтернатива находится не на первом месте.
More than 80\% of all corrected errors belong to the target class (iii).
%Главным достижением этой работы является то, что благодаря переоценке 15 альтернатив удалось сократить число ошибок iii класса на 35\%.
The main achievement of the proposed algorithm is that by combined score employment we managed to reduce the number of class (iii) errors by 40\% (see the fourth row in Table \ref{tab:comparison}).
%Как и предполагалось исправились ошибки (TODO: расписать подробнее) со сложным фоном (Fig. \ref{fig:resolved_errors}.a and \ref{fig:resolved_errors}.b).
As it was expected, the proposed algorithm was able to fix cases when the following two conditions were met:
(i) the quadrilateral with the highest contour score was formed by background lines and
(ii) its difference between foreground and background was small (Fig. \ref{fig:resolved_errors}.a and \ref{fig:resolved_errors}.b).
%Более того, иногда удается опередить большие альтернативы (Fig. \ref{fig:resolved_errors}.с) а также окклюзии (Fig. \ref{fig:resolved_errors}.d).
Moreover, the proposed algorithm sometimes was able to fix cases with strong background and foreground difference (Fig.~\ref{fig:resolved_errors}.c), and cases featuring the occlusion of the object~(Fig.~\ref{fig:resolved_errors}.d).

%Также был произведен замер времени работы системы на мобильном телефоне iPhone 6 в однопоточном режиме.
The runtime of the system was measured on iPhone~6 in single-thread mode.
%Для обработки 100 случайных изображений датасета MIDV500 базовой версии потребовалось 8.2 секунды, версии с комбинированной  оценкой первых 15 альтернатив 9 секунд, что на 8\% больше.
For this experiment, 100 random images from the MIDV-500 dataset were selected.
The system with only contour-based score required 82 ms per image, while the runtime of the system with the combined score was 88 ms per image, which is 7.3\% slower.

\subsection{Comparison with state-of-the-art}
\label{subsec:state-of-the-art}
%[in English]Для сравнения нашего алгоритма на стенде MIDV-500 с state-of-the-art алгоритмами по детекции документа мы проэкзаминовали на наличие открытого кода 5 лучших систем по качеству распознавания на датасете SmartDoc.
To compare our method with state-of-the-art document detection algorithms using the MIDV-500 dataset, we looked for the open source code of five best systems according to their performance on SmartDoc dataset.
%[in English]Оказалось, что только у одной из них имеется открытй код:
It turned out that only one of them is an open source algorithm:
%%For comparison on MIDV-500, we have tested two algorithms with an open access code.
%Both of them are based on neural networks. 
in \cite{javed2017real}, the recursive CNN for corner detection was applied. 
Their pretrained model which we used in this experiment was optimized on SmartDoc dataset, so we were able to validate the results of our measurements.
%[in English]Также мы обнаружили еще одну работу \cite{junior2020fast} с открытым кодом, нейронная сеть в которой базировалась на архитуре U-net.
We also found another work \cite{junior2020fast} with open source code, which employs neural network based on U-Net \cite{ronneberger2015u} architecture.
%The second neural network from paper \cite{junior2020fast} is based on U-Net architecture. 
The proposed model uses much less parameters than the original U-Net, and it has similar performance quality to the semantic segmentation on their synthetic private dataset. 
Unfortunately, there are no results for performance quality on an open dataset described, so we were not able to validate the correctness of our experiments on MIDV-500. 

To evaluate the performance quality, we used Jaccard index~(\ref{eq:JI}) averaged over all frames. 

Results of performance quality evaluation are demonstrated in Table \ref{tab:midv}. 
It clearly shows that our algorithm with modification allows for the highest performance quality.
Low quality of other methods can be explained by the fact that we use the pretrained models, which did not encounter the documents and backgrounds from MIDV-500.

Also, we compare our algorithm with the system from \cite{sheshkus2020houghencoder}.
It uses U-Net-like neural network with Fast Hough Transform layers.
And unlike other works, their model was trained on the subset of MIDV-500, where at least 3 vertices are within the frame. 
Only in this experiment, for the comparison we had to measure our algorithm quality performance by the average value over the subset of another variation of Jaccard index:
\begin{equation}
\frac{1}{2} \left (\frac{\mathrm{area}(M(q)\cap M(m))}{\mathrm{area}(M(q)\cup M(m))} + \frac{\mathrm{area}(\overline{M(q)}\cap \overline{M(m)})}{\mathrm{area}(\overline{M(q)}\cup \overline{M(m)})} \right),
\label{eq:jaccard_sym}
\end{equation}
where $M(.)$ is a binary mask with the same size as the input image~$I$.
This metric takes into account not only the overlay of the object, but also the background.
The algorithm from \cite{sheshkus2020houghencoder} reached 0.96 in metric (\ref{eq:jaccard_sym}), while our algorithm with contrast-based score reached 0.974 (Table \ref{tab:midv}).
Thus, the proposed method provides unmatched state-of-the-art performance on the open MIDV-500 dataset.

\begin{table}[h]
\caption{Comparison of quadrilateral detection on MIDV-500. The results in the bottom part of the table are calculated by  (\ref{eq:jaccard_sym})}
\label{tab:midv}
\begin{tabular}{cccc}
\hline
\hline
Algorithm & \begin{tabular}[c]{@{}c@{}}MIDV-500\\ 4 vertices  in\end{tabular} & \begin{tabular}[c]{@{}c@{}}MIDV-500\\ at least 3 in\end{tabular} & \begin{tabular}[c]{@{}c@{}}MIDV-500\\ full\end{tabular} \\ \hline
\hline
{[}Our{]}: Contour & 0.968 & 0.955 & 0.861 \\ 
{[}Our{]}: Combined & \textbf{0.972} & \textbf{0.961} & \textbf{0.87} \\ 
CS-NUST-2 \cite{javed2017real} & 0.739 & 0.705 & 0.626 \\ 
OctHU-PageScan \cite{junior2020fast} & 0.403 & 0.374 & 0.319 \\ 
\hline
{[}Our{]}: Combined & - & \textbf{0.974}$^*$ & - \\ 
HoughEncoder \cite{sheshkus2020houghencoder} & - & 0.96$^*$ & - \\
\hline
\hline
\end{tabular}
\end{table}

\begin{table*}[ht]
\caption{Comparison with state-of-the-art on SmartDoc.}
\label{tab:smartdoc}
\centering
\begin{tabular}{ccccccc}
\hline
\hline
Algorithm & Background 1 & Background 2 & Background 3 & Background 4 & Background 5 & All \\
\hline
\hline
{[}Our{]}: Contour & 0.980 & 0.974 & 0.982 & 0.966 & 0.294 & 0.906 \\
{[}Our{]}: Combined & 0.983 & 0.974 & 0.983 & 0.970 & 0.319 & 0.910 \\
CS-NUST-2 \cite{javed2017real} & 0.988 & 0.976 & 0.984 & 0.974 & 0.948 & 0.978 \\
JCD+CSR \cite{zhu2019coarse} & 0.988 & \textbf{0.984} & 0.983 & \textbf{0.984} & \textbf{0.961} & \textbf{0.982} \\
GOP \cite{leal2016smartphone} & 0.961 & 0.944 & 0.965 & 0.930 & 0.412 & 0.896 \\
LRDE-2 \cite{puybareau2018real} & 0.905 & 0.936 & 0.859 & 0.903 & - & - \\
LRDE-3 \cite{ngoc2019document} & 0.985 & 0.982 & 0.987 & 0.980 & 0.848 & 0.970 \\
\begin{tabular}[c]{@{}c@{}}DBSCAN \cite{el2018document}\end{tabular} & - & - & - & - & - & 0.942 \\
Smart Engines \cite{zhukovsky2017segments} & \textbf{0.989} & 0.983 & \textbf{0.990} & 0.979 & 0.688 & 0.955 \\
SmartDoc (Averaged) \cite{burie2015icdar2015} & 0.947 & 0.903 & 0.938 & 0.812 & 0.404 & 0.855 \\
\hline
\hline
\end{tabular}
\end{table*}

\begin{figure}[htb]

\footnotesize
\begin{minipage}[b]{0.98\linewidth}
  \centering
  \centerline{\includegraphics[width=\textwidth]{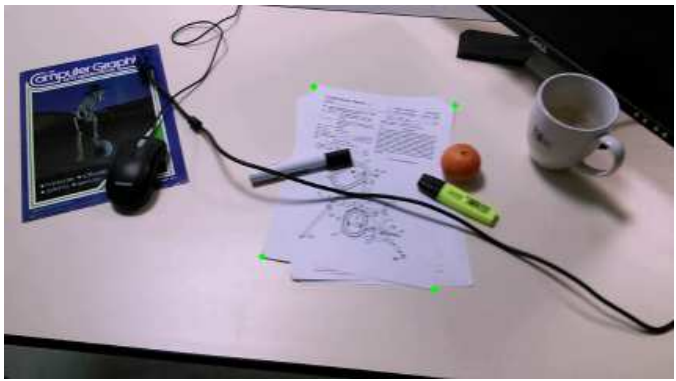}}
%  \vspace{1.5cm}
\end{minipage}

\caption{Sample image from the fifth background of SmartDoc \cite{burie2015icdar2015} featuring document to be detected highlighted by 4 green dots.}
\label{fig:5_backgound}

\end{figure}

Let us consider the proposed algorithm performance on SmartDoc \cite{burie2015icdar2015} dataset as well as all the available statistics for algorithms with open published papers listed in Table~\ref{tab:smartdoc}.
This table clearly shows that both of our algorithms demonstrate competitive results on top 4 backgrounds. 
The main cause of their failure on the fifth background is associated with the fact that our implementation of the quadrilateral detection algorithm is based on contour approach.
Since only short parts of the original quadrilateral sides have non-zero contrast, our implementation of the line detection algorithm based on FHT transform often is not able to find all sides of the original quadrilateral.
The second reason of failure is associated with the presence of other quadrilateral object (Fig.~\ref{fig:5_backgound}) within the frame featuring much stronger sides and much higher contrast between internal and external regions of the document boundaries. 

\section{Discussion}
\label{sec:discussion}

%[TODO]В предыдущей главе было продемонстрировано, что предложенный алгоритм показывает state-of-the-art качество на датасете MIDV-500 а также высокое качество детекции документа на большинстве бэкграундов стенда SmartDoc.
In the previous section, we demonstrated that the proposed algorithm allows for state-of-the-art performance quality for the MIDV-500 dataset, and a high document detection quality for the majority of the SmartDoc backgrounds.
%[TODO]Однако, анализ смартодока на первый взгляд показывает, что комбенированный подход не дает ожидаемого улучшения.
Nevertheless, the preliminary analysis of the SmartDoc results shows that the combined approach does not yield the expected improvement.
%[TODO]Но при детальном раммотрении видно, что в первых четырех случаях чинить с помощью комбинированного подхода нечего, поскольку уже хорошо работает алгоритм, использующий только котурную оценку. Комбинированная оценка при этом дает стабильно небльшой прирост. 
But a more detailed examination demonstrates that for the first four SmartDoc backgrounds, there are no image distortions which require the combined approach, since contour-based algorithm works well for the images included in these sets.  
%[TODO]Случай пятого фона заслуживает отдельного рассморения.
But the fifth background is a subject of a separate discussion. 

Since the fifth background of the SmartDoc dataset includes images capturing two quadrilateral objects, target A4 document and a mouse pad (Fig.~\ref{fig:5_backgound}), and the latter features much stronger sides and much higher contrast between internal and external regions of the document boundaries, thus, the detected quadrilateral often corresponds to the mouse pad.
Such behavior of the system does not contradict our problem statement, because by problem statement there must be only one document within the frame and more than that information about document internal structure is unknown.
Therefore, in the fifth background of SmartDoc, we re-annotated the images which contain two quadrilateral objects.
In such cases, we marked as a ground truth the object with maximum areal contrast.
After this re-annotation, we measured the performance quality on the fifth background again.
The accuracy of the algorithm with only contour-based score was 0.407, and with the combined score -- 0.478.
%[TODO]Итак, если учитывать нашу постановку задачи, aлгоритм отработал не так плохо, поскольку прирост от использования комбинированной оценки увеличился.
Thus, given our problem statement, the performance quality of the proposed algorithm is not bad, since combined score allowed for a higher accuracy.
%[TODO]Тем не менее такой результат для алгоритма, использующего комбинированную оценку, нельзя признать удовлетворительным.
But this result is not sufficient for the combined approach.

In our modification, we use the feature employed in region-based approaches only for the ranking of the quadrilaterals, which were constructed by the contour-based algorithm.
This modification is not able to detect the original quadrilateral, if the latter did not get into the list of quadrilaterals to be ranked.
Such outcomes are demonstrated in fifth background of SmartDoc dataset, where the performance quality of our contour-based algorithm implementation is low. 
To check the modification performance within the fifth background, we designed the following experiment for the algorithm employing combined score.
Let us consider the alternative with the lowest contour-based score of all the alternatives in the ranking list.
We replaced this alternative with the ground truth quadrilateral, and ran the experiments featuring the fifth background again. 
The quadrilateral detection quality in this case increases from 0.319 to 0.751 (and from 0.478 to 0.917 in case of the re-annotated fifth background of SmartDoc), which demonstrates the applicability of the suggested modification. 
%[TODO]Позвольте заметить, что качество 0.751 (0.917 for re-marked subset) достигается перевзвешиванием верной альтернативы, если бы она имела 11 вес, среди ранжируемых альтернатив.
Note that the quality of 0.751 (0.917 for re-annotated subset) is achieved when the correct alternative is assigned to be the 11th alternative in the ranking list.

%[TODO]Как было сказано выше, качество алгоритма, используемого только контурную оценку, достаточно низкое.
As mentioned before, the quality of the contour-based algorithm is pretty low.
%[TODO]Это обусловлено тем, что контраста у границ почти нет, следовательно нам не удается найти step-edges, которые мы используем для выделения контуров.
This is because the borders lack contrast, thus we cannot find the step-edges used in contour detection.
%[TODO]Возможно, результаты бы стали лушче, если использовать roof-edges (ridges), как было сделано в \cite{tropin2019localization}.
Perhaps the results would have been better, if the roof-edges (ridges) were used, as in \cite{tropin2019localization}.
%[TODO]Однако, при использовани roof-edges нельзя (можно ослабить до "опасно") применять даунскейл так, как мы его используем при детектировании step-edge: step-edges are scale-invariant because the sizes of the document are much bigger than the scale factor \cite{canny1986computational}.
But in the case of roof-edges employment, it is dangerous to use downscale as we did in step-edge detection (step-edges are scale-invariant because the size of the document is much bigger than the used scale factor~\cite{canny1986computational} and it is not true for the roof-edges).
%[TODO]For roof-edges it is not true.

For our modification it does not matter how exactly the set of quadrilateral alternatives to be ranked was generated.
In this work, we did not consider geometric properties of the object while baseline set of quadrilaterals generation, since we are focused primarily on the contour-based score applicability to the quadrilateral search.

Also we would like to add that unlike the algorithm suggested in \cite{javed2017real}, which demonstrates one of the highest performance quality on SmartDoc dataset, but needs training for the MIDV-500 dataset, our algorithm does not require any additional tuning. 

\section{Conclusion}
\label{sec:conclusion}
In this paper, the improvement of the classical contour-based approach for the document detection was proposed.
The scoring function for the ranking of the quadrilateral candidates was modified: it takes into account not only contour characteristics, but also the degree of difference between internal and external regions of the candidate.
This modification reduced the number of the ranking errors by 40\%, while retaining its applicability on the mobile phones.
The proposed algorithm was also tested on the open datasets.
The suggested algorithm demonstrated the highest quality on MIDV-500 dataset and the competitive results on four out of five parts of SmartDoc dataset.

\section{Future work}
To increase the performance quality of our algorithm in the future, we are going to take into account the geometric properties of the document and its inter-frame transformation in the video stream.

\bibliographystyle{IEEEtran}
\bibliography{IEEEabrv,mybibfile}

\begin{thebibliography}{10}
\providecommand{\url}[1]{#1}
\csname url@samestyle\endcsname
\providecommand{\newblock}{\relax}
\providecommand{\bibinfo}[2]{#2}
\providecommand{\BIBentrySTDinterwordspacing}{\spaceskip=0pt\relax}
\providecommand{\BIBentryALTinterwordstretchfactor}{4}
\providecommand{\BIBentryALTinterwordspacing}{\spaceskip=\fontdimen2\font plus
\BIBentryALTinterwordstretchfactor\fontdimen3\font minus
  \fontdimen4\font\relax}
\providecommand{\BIBforeignlanguage}[2]{{%
\expandafter\ifx\csname l@#1\endcsname\relax
\typeout{** WARNING: IEEEtran.bst: No hyphenation pattern has been}%
\typeout{** loaded for the language `#1'. Using the pattern for}%
\typeout{** the default language instead.}%
\else
\language=\csname l@#1\endcsname
\fi
#2}}
\providecommand{\BIBdecl}{\relax}
\BIBdecl

\bibitem{attivissimo2019automatic}
F.~Attivissimo, N.~Giaquinto, M.~Scarpetta, and M.~Spadavecchia, ``An automatic
  reader of identity documents,'' in \emph{2019 IEEE International Conference
  on Systems, Man and Cybernetics (SMC)}.\hskip 1em plus 0.5em minus
  0.4em\relax IEEE, 2019, pp. 3525--3530.

\bibitem{tam2003quadrilateral}
A.~Tam, H.~Shen, J.~Liu, and X.~Tang, ``Quadrilateral signboard detection and
  text extraction.'' in \emph{CISST}, 2003, pp. 708--713.

\bibitem{zhang2007whiteboard}
Z.~Zhang and L.-W. He, ``Whiteboard scanning and image enhancement,''
  \emph{Digital Signal Processing}, vol.~17, no.~2, pp. 414--432, 2007.

\bibitem{duan2004combining}
T.~D. Duan, D.~A. Duc, and T.~L.~H. Du, ``Combining hough transform and contour
  algorithm for detecting vehicles' license-plates,'' in \emph{Proceedings of
  2004 International Symposium on Intelligent Multimedia, Video and Speech
  Processing, 2004.}\hskip 1em plus 0.5em minus 0.4em\relax IEEE, 2004, pp.
  747--750.

\bibitem{wenzel2017corners}
T.~Wenzel, T.-W. Chou, S.~Brueggert, and J.~Denzler, ``From corners to
  rectangles—directional road sign detection using learned corner
  representations,'' in \emph{2017 IEEE Intelligent Vehicles Symposium
  (IV)}.\hskip 1em plus 0.5em minus 0.4em\relax IEEE, 2017, pp. 1039--1044.

\bibitem{skoryukina20192d}
N.~S. Skoryukina, D.~P. Nikolaev, and V.~V. Arlazarov, ``2d art recognition in
  uncontrolled conditions using one-shot learning,'' in \emph{Eleventh
  International Conference on Machine Vision (ICMV 2018)}, vol. 11041.\hskip
  1em plus 0.5em minus 0.4em\relax International Society for Optics and
  Photonics, 2019, p. 110412E.

\bibitem{kittipanya2012bed}
P.~Kittipanya-Ngam, O.~S. Guat, and E.~H. Lung, ``Bed detection for monitoring
  system in hospital wards,'' in \emph{2012 Annual International Conference of
  the IEEE Engineering in Medicine and Biology Society}.\hskip 1em plus 0.5em
  minus 0.4em\relax IEEE, 2012, pp. 5887--5890.

\bibitem{sark2019artractor}
A.~Sarkar, L.~Stowe, and A.~Petersen, ``Tractor trailer bsm parameters
  estimation for smart tractor v2v deployment using cameras,'' in
  \emph{Proceedings of the 5th International Symposium on Future Active Safety
  Technology toward Zero Accidents}, 2019.

\bibitem{peter2011using}
M.~Peter, N.~Haala, and D.~Fritsch, ``Using photographed evacuation plans to
  support mems imu navigation,'' in \emph{Proceedings of the 2011 International
  Conference on Indoor Positioning and Indoor Navigation (IPIN2011), Guimaraes,
  Portugal}, 2011, pp. 30--81.

\bibitem{arlazarov2019midv}
V.~V. Arlazarov, K.~B. Bulatov, T.~S. Chernov, and V.~L. Arlazarov, ``Midv-500:
  a dataset for identity document analysis and recognition on mobile devices in
  video stream,'' \emph{Computer optics}, vol.~43, no.~5, 2019.

\bibitem{liu2018dynamic}
N.~Liu and L.~Wang, ``Dynamic detection of an object framework in a mobile
  device captured image,'' Nov.~20 2018, uS Patent 10,134,163.

\bibitem{skoryukina2015real}
N.~Skoryukina, D.~P. Nikolaev, A.~Sheshkus, and D.~Polevoy, ``Real time
  rectangular document detection on mobile devices,'' in \emph{Seventh
  International Conference on Machine Vision (ICMV 2014)}, vol. 9445.\hskip 1em
  plus 0.5em minus 0.4em\relax International Society for Optics and Photonics,
  2015, p. 94452A.

\bibitem{lampert2005oblivious}
C.~H. Lampert, T.~Braun, A.~Ulges, D.~Keysers, and T.~M. Breuel, ``Oblivious
  document capture and real-time retrieval,'' in \emph{International Workshop
  on Camera Based Document Analysis and Recognition (CBDAR)}, vol.~8.\hskip 1em
  plus 0.5em minus 0.4em\relax Citeseer, 2005.

\bibitem{hirzer2008marker}
M.~Hirzer, ``Marker detection for augmented reality applications,'' in
  \emph{Seminar/Project Image Analysis Graz}, vol.~25, 2008.

\bibitem{awal2017complex}
A.~M. Awal, N.~Ghanmi, R.~Sicre, and T.~Furon, ``Complex document
  classification and localization application on identity document images,'' in
  \emph{2017 14th IAPR International Conference on Document Analysis and
  Recognition (ICDAR)}, vol.~1.\hskip 1em plus 0.5em minus 0.4em\relax IEEE,
  2017, pp. 426--431.

\bibitem{fan2016detection}
J.~Fan, ``Detection of quadrilateral document regions from digital
  photographs,'' in \emph{2016 IEEE Winter Conference on Applications of
  Computer Vision (WACV)}.\hskip 1em plus 0.5em minus 0.4em\relax IEEE, 2016,
  pp. 1--9.

\bibitem{usilin2010visual}
S.~Usilin, D.~Nikolaev, V.~Postnikov, and G.~Schaefer, ``Visual appearance
  based document image classification,'' in \emph{2010 IEEE International
  Conference on Image Processing}.\hskip 1em plus 0.5em minus 0.4em\relax IEEE,
  2010, pp. 2133--2136.

\bibitem{bohush2019video}
R.~Bohush, A.~Kurilovich, and S.~Ablameyko, ``Video-based content extraction
  algorithm from bank cards for ios mobile devices,'' in \emph{International
  Conference on Pattern Recognition and Information Processing}.\hskip 1em plus
  0.5em minus 0.4em\relax Springer, 2019, pp. 180--191.

\bibitem{puybareau2018real}
{\'E}.~Puybareau and T.~G{\'e}raud, ``Real-time document detection in
  smartphone videos,'' in \emph{2018 25th IEEE International Conference on
  Image Processing (ICIP)}.\hskip 1em plus 0.5em minus 0.4em\relax IEEE, 2018,
  pp. 1498--1502.

\bibitem{ngoc2019document}
M.~{\^O}.~V. Ngoc, J.~Fabrizio, and T.~G{\'e}raud, ``Document detection in
  videos captured by smartphones using a saliency-based method,'' in \emph{2019
  International Conference on Document Analysis and Recognition Workshops
  (ICDARW)}, vol.~4.\hskip 1em plus 0.5em minus 0.4em\relax IEEE, 2019, pp.
  19--24.

\bibitem{leal2016smartphone}
L.~R. Leal and B.~L. Bezerra, ``Smartphone camera document detection via
  geodesic object proposals,'' in \emph{2016 IEEE Latin American Conference on
  Computational Intelligence (LA-CCI)}.\hskip 1em plus 0.5em minus 0.4em\relax
  IEEE, 2016, pp. 1--6.

\bibitem{javed2017real}
K.~Javed and F.~Shafait, ``Real-time document localization in natural images by
  recursive application of a cnn,'' in \emph{2017 14th IAPR International
  Conference on Document Analysis and Recognition (ICDAR)}, vol.~1.\hskip 1em
  plus 0.5em minus 0.4em\relax IEEE, 2017, pp. 105--110.

\bibitem{zhu2019coarse}
A.~Zhu, C.~Zhang, Z.~Li, and S.~Xiong, ``Coarse-to-fine document localization
  in natural scene image with regional attention and recursive corner
  refinement,'' \emph{International Journal on Document Analysis and
  Recognition (IJDAR)}, vol.~22, no.~3, pp. 351--360, 2019.

\bibitem{zhukovsky2017segments}
A.~Zhukovsky, D.~Nikolaev, V.~Arlazarov, V.~Postnikov, D.~Polevoy,
  N.~Skoryukina, T.~Chernov, J.~Shemiakina, A.~Mukovozov, I.~Konovalenko
  \emph{et~al.}, ``Segments graph-based approach for document capture in a
  smartphone video stream,'' in \emph{2017 14th IAPR International Conference
  on Document Analysis and Recognition (ICDAR)}, vol.~1.\hskip 1em plus 0.5em
  minus 0.4em\relax IEEE, 2017, pp. 337--342.

\bibitem{junior2020fast}
R.~B. d.~N. Junior, L.~F. Ver{\c{c}}osa, D.~Mac{\^e}do, B.~L.~D. Bezerra, and
  C.~Zanchettin, ``A fast fully octave convolutional neural network for
  document image segmentation,'' \emph{arXiv preprint arXiv:2004.01317}, 2020.

\bibitem{sheshkus2020houghencoder}
\BIBentryALTinterwordspacing
A.~Sheshkus, D.~Nikolaev, and V.~L. Arlazarov, ``Houghencoder: neural network
  architecture for document image semantic segmentation,'' \emph{preprint},
  2020. [Online]. Available:
  \url{ftp://smartengines.com/preprints/sheshkus2020houghencoder.pdf}
\BIBentrySTDinterwordspacing

\bibitem{castelblanco2020machine}
A.~Castelblanco, J.~Solano, C.~Lopez, E.~Rivera, L.~Tengana, and M.~Ochoa,
  ``Machine learning techniques for identity document verification in
  uncontrolled environments: A case study,'' in \emph{Mexican Conference on
  Pattern Recognition}.\hskip 1em plus 0.5em minus 0.4em\relax Springer, 2020,
  pp. 271--281.

\bibitem{canny1986computational}
J.~Canny, ``A computational approach to edge detection,'' \emph{IEEE
  Transactions on pattern analysis and machine intelligence}, no.~6, pp.
  679--698, 1986.

\bibitem{bobkov2006matching}
V.~Bobkov, Y.~Ronshin, and A.~Kudrashov, ``Sopostavlenie linij po trem vidam
  prostranstvennoj sceny,'' \emph{Journal of information technologies and
  computing systems}, no.~2, pp. 71--78, 2006.

\bibitem{brady1998fast}
M.~L. Brady, ``A fast discrete approximation algorithm for the radon
  transform,'' \emph{SIAM Journal on Computing}, vol.~27, no.~1, pp. 107--119,
  1998.

\bibitem{burie2015icdar2015}
J.-C. Burie, J.~Chazalon, M.~Coustaty, S.~Eskenazi, M.~M. Luqman, M.~Mehri,
  N.~Nayef, J.-M. Ogier, S.~Prum, and M.~Rusi{\~n}ol, ``Icdar2015 competition
  on smartphone document capture and ocr (smartdoc),'' in \emph{2015 13th
  International Conference on Document Analysis and Recognition (ICDAR)}.\hskip
  1em plus 0.5em minus 0.4em\relax IEEE, 2015, pp. 1161--1165.

\bibitem{ronneberger2015u}
O.~Ronneberger, P.~Fischer, and T.~Brox, ``U-net: Convolutional networks for
  biomedical image segmentation,'' in \emph{International Conference on Medical
  image computing and computer-assisted intervention}.\hskip 1em plus 0.5em
  minus 0.4em\relax Springer, 2015, pp. 234--241.

\bibitem{el2018document}
H.~El~Bahi and A.~Zatni, ``Document text detection in video frames acquired by
  a smartphone based on line segment detector and dbscan clustering,''
  \emph{Journal of Engineering Science and Technology}, vol.~13, no.~2, pp.
  540--557, 2018.

\bibitem{tropin2019localization}
D.~Tropin, J.~Shemiakina, I.~Konovalenko, and I.~Faradjev, ``Localization of
  planar objects on the images with complex structure of projective
  distortion,'' \emph{Information processes}, vol.~19, no.~2, pp. 208--229,
  2019.

\end{thebibliography}

\end{document}